\documentclass{article}

\usepackage{microtype}
\usepackage{graphicx}
\usepackage{subfigure}
\usepackage{booktabs} 
\usepackage{caption}

\usepackage{hyperref}

\usepackage[accepted]{icml2023}

\usepackage{amsmath}
\usepackage{amssymb}
\usepackage{mathtools}
\usepackage{amsthm}
\usepackage{adjustbox} 
\usepackage[capitalize,noabbrev]{cleveref}

\theoremstyle{plain}

\theoremstyle{definition}

\theoremstyle{remark}

\usepackage[textsize=tiny]{todonotes}

\newcommand{\algname}{MFCL}
\newcommand{\loss}{\mathcal{L}}
\newcommand{\model}{\mathcal{F}}
\newcommand{\gen}{\mathcal{G}}
\newcommand{\normal}{\mathcal{N}}

\newcommand{\synX}{\Tilde{x}}

\newcommand{\info}{\mathcal{H}_{info}}
\newcommand{\weight}{\mathcal{W}}

\icmltitlerunning{Don’t Memorize; Mimic The Past: Federated Class Incremental Learning Without Episodic Memory}

\begin{document}

\twocolumn[
\icmltitle{Don’t Memorize; Mimic The Past: Federated Class Incremental Learning Without Episodic Memory}



\begin{icmlauthorlist}
\icmlauthor{Sara Babakniya}{cs}
\icmlauthor{Zalan Fabian}{ee}
\icmlauthor{Chaoyang He}{comp}
\icmlauthor{Mahdi Soltanolkotabi}{ee}
\icmlauthor{Salman Avestimehr}{ee}
\end{icmlauthorlist}

\icmlaffiliation{cs}{Department of Computer Science, University of Southern California, Los Angeles, USA}
\icmlaffiliation{ee}{Ming Hsieh Department of Electrical Engineering, University of Southern California, Los Angeles, USA}
\icmlaffiliation{comp}{FedML}

\icmlcorrespondingauthor{Sara Babakniya}{babakniy@usc.edu}

\icmlkeywords{Federated Learning, ICML, Continual Learning}

\vskip 0.3in
]



\printAffiliationsAndNotice{} 

\begin{abstract}
Deep learning models are prone to forgetting information learned in the past when trained on new data. This problem becomes even more pronounced in the context of federated learning (FL), where data is decentralized and subject to independent changes for each user. Continual Learning (CL) studies this so-called \textit{catastrophic forgetting} phenomenon primarily in centralized settings, where the learner has direct access to the complete training dataset. However, applying CL techniques to FL is not straightforward due to privacy concerns and resource limitations. 
This paper presents a framework for federated class incremental learning that utilizes a generative model to synthesize samples from past distributions instead of storing part of past data. Then, clients can leverage the generative model to mitigate catastrophic forgetting locally. The generative model is trained on the server using data-free methods at the end of each task without requesting data from clients. Therefore, it reduces the risk of data leakage as opposed to training it on the client's private data. We demonstrate significant improvements for the CIFAR-100 dataset compared to existing baselines.
\end{abstract}




\section{Introduction}
\label{sec:intro}
Federated learning (FL) \cite{mcmahan2017communication, konevcny2016federated}
is a decentralized machine learning technique that enables privacy-preserving collaborative learning. In FL, multiple users (clients) train a common (global) model in coordination with a centralized node (server) without sharing personal data. In recent years, FL has attracted tremendous attention in both research and industry and has been successfully employed in various fields. 
Despite its popularity, deploying FL in practice requires addressing challenges such as resource limitation and statistical  heterogeneity \cite{kairouz2021advances}. Furthermore, there are still common assumptions in most FL frameworks that are far from reality.
One such assumption is that the client's local data distribution does not change over time. However, in real-world \cite{shoham2019overcoming}, users' data constantly evolve due to changes in the environment or trends. In such scenarios, the model must rapidly adapt to the incoming data while preserving performance in the past. 

In the centralized setting, such problems have been explored in continual learning \cite{shin2017continual,li2017learning}. Despite all the significant progress for the centralized problems, most methods cannot be directly employed in the FL setting due to inherent differences between the two settings. For instance, experience replay \cite{rolnick2019experience} is a popular approach, where a portion of past data points is saved to maintain some representation of past distributions throughout the training. However, deploying experience replay in FL has resource and privacy limitations. It requires clients to store and keep their data which may not be possible because of privacy reasons. This can be highly important, especially in cases for example a service provider can store its customer's data for only a short time. Besides, even storing is possible; such data overhead increase the memory usage of already resource-limited clients. 

To address the above problems, we propose \algname, \emph{\underline{M}imicking \underline{F}ederated \underline{C}ontinual \underline{L}earning}. In particular, \algname{} is based on training a generative model in the server and sharing it with clients to sample synthetic examples of past data instead of clients storing their data. The generative model training is data-free in the sense that it only requires the global model without any form of training data from the clients. This is specifically important because this step does not require powerful clients and does not cause any extra data leakage from them. Finally, our experiments demonstrate improvement by $20\%$ in average accuracy while reducing the training overhead of the clients.

We summarize our contributions below:
\begin{itemize}
\setlength\itemsep{-0.2em}
    \item We propose a novel framework to tackle the problem of federated class incremental learning more efficiently. Our framework specifically targets applications where past samples are unavailable.
    \item We point out potential issues with relying on client-side memory for FCL, and propose using a generative model trained by the server in a \textit{data-free manner} to reduce catastrophic forgetting while preserving privacy.
    \item We demonstrate the efficacy of our method in more realistic scenarios with a larger number of clients and a more challenging dataset (CIFAR-100). 
\end{itemize}

\section{Related Work}
\textbf{Continual Learning.} Catastrophic forgetting \cite{mccloskey1989catastrophic} is a fundamental problem: when we train a model on new examples, its performance on past data degrades. This problem is investigated in continual learning \cite{zenke2017continual}, and the goal is for the model to learn new information while preserving its old ones.

Recent works focus on three scenarios, namely task-, domain- and class-incremental learning \cite{van2019three}. In \textit{Task-IL}, tasks are disjoint, and the output spaces are separated by task IDs provided during training and test time. For \textit{Domain-IL}, the output space is still the same, but the task IDs are no more provided. Finally, in \textit{Class-IL}, new tasks introduce new classes to the output space, and the number of classes increases incrementally. Among these scenarios, we focus on \textbf{Class-IL}, which is the more challenging and realistic, especially in FL. In the FL applications, there is no task ID available, and it is preferred to learn a \textit{single} model useable for all the observed data.

\textbf{Federated Continual Learning.} In Federated Continual Learning (FCL), the main focus is to adapt the global model to new data while maintaining knowledge of past data, all under the standard restrictions of FL. This important problem has only gained attention very recently, and \cite{yoon2021federated} is the first paper on this topic. It focuses on Task-IL, which requires a unique task id per task during inference. Furthermore, it adapts separate masks per task to improve personalized performance without preserving a common global model. This setting is considerably different than ours as we target class-IL with a single global model that can classify all the classes seen so far. \cite{macontinual} employs knowledge distillation using a surrogate dataset. \cite{dong2022federated} relaxes the problem as clients have access to large memory to save the old examples and share their data which is different from the standard FL setting. \cite{jiang2021fedspeech,priyanshu2021continual,usmanova2021distillation} explore the FCL problem in domains other than image classification.

This work focuses on Class-IL for supervised image classification without memory replay, which has been also discussed in \cite{qi2023better,hendryx2021federated}. However, \cite{hendryx2021federated} allows overlapping classes between tasks and focuses on few-shot learning, which is different from the standard class-IL. The most closely related work to ours is \cite{qi2023better}, where authors propose FedCIL. This work also benefits from methods based on generative replay to compensate for the absence of old data and overcome forgetting. In FedCIL, clients train the discriminator and generator locally. Then, the server takes a consolidation step after aggregating the updates. In this step, the server generates synthetic data using all the generative models trained by the clients to consolidate the global model and improve the performance. The main difference between this work and ours is that in our work, the generative model is trained by the server in a data-free manner which can reduce clients' computation and does not require their private data. 


\textbf{Data-Free Knowledge Distillation.} Knowledge distillation (KD)\cite{hinton2015distilling} is a popular method to transfer knowledge from a well-trained teacher model to a (usually) smaller student model using at least a small portion of training data. However, in cases that such data is unavailable (e,g, privacy concerns), a new line of work \cite{chen2019data,haroush2020knowledge} proposes \textit{data-free knowledge distillation}. In such methods, a generative model is used as a training data substitute. This model generates synthetic data such that the teacher model predicts them as their assigned label. Data-free KD has been previously used in FL \cite{zhu2021data} as a solution for data heterogeneity. However, to the best of our knowledge, this is the first work that adapted such a technique in the context of FCL. 
\begin{figure*}[t]
     \centering
     \subfigure{
         \centering
         \includegraphics[width=0.45\linewidth]{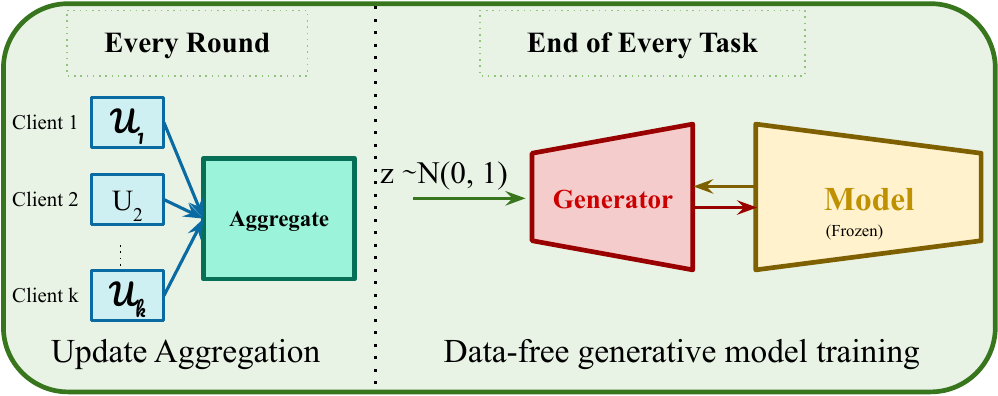}
     }
      \centering
     \subfigure{
        \centering
         \includegraphics[width=0.45\linewidth]{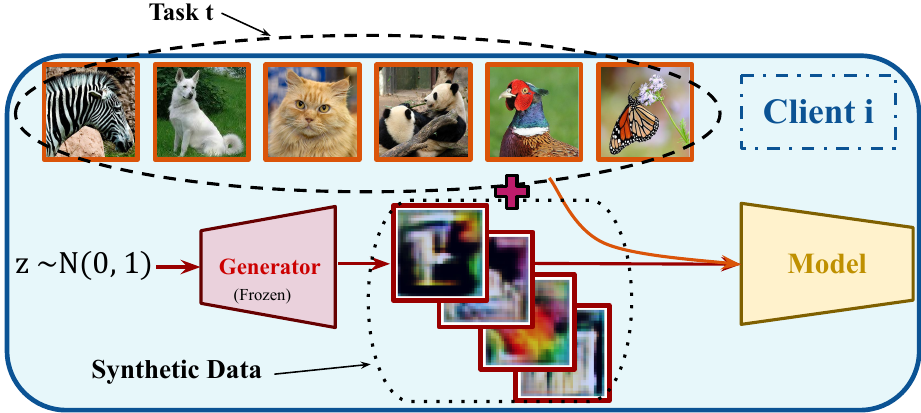}
     }
    \caption{[Left] Aggregation (every round), generator training (end of each task). [Right] Clients train models using synthetic + local data.}
    \label{fig:alg}
\end{figure*}
\section{Federated Class-IL with \algname}
In federated class-IL, a shared model is trained on $T$ tasks. However, the distributed nature of FL makes it distinct from the centralized version. In FL, users may join, drop out or change their data independently. Also, required data or computation power for some centralized algorithms may not be available due to privacy and resource constraints.

To address the mentioned problems, we propose \algname. This algorithm includes two essential parts: \textit{first}, at the end of each task, the server trains a generative model with data-free knowledge distillation methods to learn the representation of the seen classes. \textit{Second}, clients diminish catastrophic forgetting by generating synthetic images from the generative model. This way, clients do not require memories to store old data. Also, since the server trains the generative model training without additional information, this step does not introduce new privacy issues. Finally, \algname{} can help mitigate the data heterogeneity problem, as clients can synthesize samples from classes they do not own. Here, we explain the two key parts of our algorithm: server-side  (Fig. \ref{fig:alg}. left) and client-side (Fig. \ref{fig:alg}. right).
\subsection{Server-Side: Generative Model}
The motivation for deploying a generative model is to synthesize images that mimic the old tasks and to avoid storing past data. However, training these generators on the client's side, where the training data exists, is \textit{computationally expensive} and \textit{requires a large amount of training data} and can be potentially \textit{privacy concerning}. On the other hand, the server has only access to the global model and no data. We propose training a generative model on the server, but in a data-free manner, i.e., by means of model-inversion image synthesis \cite{yin2020dreaming,smith2021always}. In such approaches, the goal is to synthesize images optimized with respect to the discriminator (global model). Then, the generative model is shared with the clients to be later used in sampling images during local training. To this aim, we utilize a generative model, $\gen$, that takes noise $z \sim \normal(0, \mathbf{I})$ as input and produces a synthetic sample $\synX$. In training this model, we employ the following training objectives.

\textbf{Cross Entropy Loss.} First, the synthetic data should be labeled correctly by the current discriminator model (global model or $\model$). Therefore, we employ cross entropy classification loss between its assigned label $z$ and the prediction of $\model$ on synthetic data $\synX$. Note, that noise dimension can be arbitrary and greater than the current discovered classes of task $t$, and we only consider the first $q$ dimension here, where $q = \sum_{i=1}^{t} |\mathcal{Y}^{i}|$. Then, we can define this loss as
\begin{align}
    \loss_{CE} = CE(argmax(z[:q]), \model(\synX)).
\end{align}

\textbf{Diversity Loss.} Synthesized images can suffer from a lack of class diversity, and we utilize information entropy (IE) \cite{chen2019data} to solve this. For a probability vector $\texttt{p}=(p_1, p_2,..., p_q)$, IE is evaluated as $\info(\texttt{p}) = - \frac{1}{q}\sum_{i}p_i \log(p_i)$. Therefore, diversity loss is defined as
\begin{align}
    \loss_{div} = -\info(\frac{1}{bs}\sum_{i=1}^{bs}\model(\synX_{i})). 
\end{align}

This loss measures the IE for samples of a batch (batch size =$bs$). Maximizing this term encourages the output distribution of the generator to be balanced for all the classes.

\begin{table*}[h!]
	\tiny\addtolength{\tabcolsep}{-2.5pt}
		\centering
 		\begin{adjustbox}{width=.99\textwidth}
        \small
		\begin{tabular}{c|c|c|c|c|c}
            & Average Accuracy & Average forgetting  &  Training time (s) &  Training time (s) & Server Runtime (s) \\
            & $\Tilde{\mathcal{A}}$ (\%)  &  $\Tilde{f}$(\%)  &  ($T=0$)&  ($T\geq 1$) & \\
			\hline
            \hline
            \textbf{FedAvg}           & $22.27\pm 0.22$          &  $78.77\pm0.83$          &  $\approx 1.2$& $\approx 1.2$ &  $\approx 1.8$  \\ 
            \textbf{FedProx}          & $22.00\pm 0.31$          &  $78.17\pm0.33$          & $\approx 1.98$& $\approx 1.98$&  $\approx 1.8$ \\ 
            \textbf{FedCIL}           & $26.8 \pm 0.44$          &  $38.19\pm0.31$          & $\approx 17.8$& $\approx 24.5$&  $\approx 2.5$ for $T=1$, $\approx 4.55$ for $T>1$  \\ 
            \textbf{FedLwF-2T}        & $22.17\pm 0.13$          &  $75.08\pm0.72$          & $\approx 1.2$ & $\approx 3.4$ &  $\approx 1.8 $ \\ 
            \textbf{\algname{} (Ours)}& $\mathbf{43.87\pm 0.12}$ & $\mathbf{28.3\pm0.78}$   & $\approx 1.2$ & $\approx 3.7$ &  $\approx 330$ (once per task), $\approx 1.8$ O.W.\\ 
            \textbf{Oracle}           & $67.12\pm 0.4 $          & $--$                     & $\approx 1.2$ & $\approx 1.2 \times\ T$  & $\approx 1.8$\\ 
		\end{tabular}
		\end{adjustbox}
  
		\caption{Evaluation on CIFAR-100 dataset.
		\label{table:results}}
		\vspace{-2mm}
\end{table*}

\textbf{Batch Statistics Loss.} Prior works \cite{haroush2020knowledge,yin2020dreaming,smith2021always} in the centralized setting have recognized that the distribution of synthetic images can drift from real data. We can use batch statistics loss $\loss_{BN}$ to avoid such problems. Specifically, the goal is to  enforce synthetic images to produce similar statistics in all BatchNorm layers to the ones that are already produced during training. To this end, we minimize the layer-wise distances between the two statistics written as
\begin{align}
    \loss_{BN} = \frac{1}{L} \sum_{i=1}^{L}KL(\normal(\mu_{i},\sigma^2_{i}), \normal(\Tilde{\mu}_{i},\Tilde{\sigma}^2_{i}))
\end{align}

 Here, $L$ denotes the number of BatchNorm layers in the model, $\mu_{i} $ and $\sigma_{i}$ are the mean and standard deviation stored in BatchNorm layer $i$ of the global model, $\Tilde{\mu}_{i},\Tilde{\sigma}_{i}$ are measured statistics of BatchNorm layer $i$ for the synthetic images, $KL$ stands for the  Kullback-Leibler divergence.

Finally, we can write the training objective of $\gen$ as \eqref{que:gen} where $w_{div}$ and $w_{BN}$ control the weight of each term.
\begin{align}
    \min_{\gen} \loss_{ce} + w_{div} \loss_{div} + w_{BN} \loss_{BN},
    \label{que:gen}
\end{align}

\subsection{Client-side: Continual Learning}
For client-side training, inspired by \cite{smith2021always}, we distill the \textit{stability-plasticity} dilemma into three critical requirements of CL and aim to address them one by one. 

\textbf{Current task.} To have plasticity, the model needs to learn the new features in a way that is least biased toward the old tasks. So, here, the CE loss is computed \textit{for the new classes only} by splitting the linear heads and excluding the old ones:
\begin{align}
    \loss_{CE}^t  = CE(\model_t(x), y) ~~ \ if\  y \in \mathcal{Y}^t\ else\ 0.
\end{align}

\textbf{Previous tasks.} To reduce forgetting, we train the model using synthetic and real data simultaneously. However, the distribution of the synthetic data differs from the real one, and it becomes important to prevent the model from distinguishing between old and new data. To address this problem, for fine-tuning the decision boundary using the sampled synthetic data ($\synX =Sample(\gen_{t-1})$), clients freeze the feature extraction part and only update the classification head (represented by $\model_t^*$). This loss can be formulated as
\begin{align}
    \loss_{FT}^t = CE(\model_t^*(\synX), y).
\end{align}

Finally, to minimize forgetting, the common method is knowledge distillation over the prediction layer. However, \cite{smith2021always} proposed \textit{importance-weighted feature distillation}: instead of using the knowledge in the decision layer, they use the output of the feature extraction part of the model (penultimate layer). This way, only the more significant features of the old model are transferred, enabling the model also to learn the new features from the new tasks. This can be written as below where $\weight$ is the frozen linear head of the model trained on the last task ($\weight = \model_{t-1}^{L}$). 
\begin{align}
    \loss_{KD}^{t} = || \weight(\model_{t}^{1: L-1}(\hat{x})) - \weight( \model_{t-1}^{1: L- 1}(\hat{x})) ||^{2}_{2},
\end{align}

In summary, the final objective on the client side as
\begin{align}
\min_{\model_t} \loss_{CE}^t + w_{FT} \loss_{FT}^t + w_{KD} \loss_{KD}^{t},
\label{que:local}
\end{align}
$w_{FT}$ and $w_{KD}$ determine the importance of each loss term.

\subsection{Algorithm $\algname$}
For the first task, clients train the model using the $\loss_{ce}$. At the end of training task $t=1$, the server trains the generative model by optimizing \eqref{que:gen}. Then, the server freezes and saves $\gen$ and the global model ($\model_{t-1}$). This procedure repeats for all future tasks, with the only difference being that for $t>1$, the server needs to send the current global model ($\model_t$), the previous task's final model ($\model_{t-1}$) and $\gen$ to clients. Since $\model_{t-1}$ and $\gen$ are fixed during the whole process of training $\model_t$, the server can send them to each client once per task to reduce the communication cost. To further decrease this overhead, we can use communication-efficient methods, such as \cite{qiu2022zerofl,babakniya2022federated}, that highly compress the model with minor performance degradation.

\section{Experiments}
\textbf{Setting.} We demonstrate the efficacy of our method on dataset: CIFAR-100 \cite{krizhevsky2009learning}. 
We use the baseline ResNet18 \cite{he2016deep} as the global model and ConvNet architecture for $\gen$. In our experiments, there are 50 clients in total and 5 randomly sampled participants in every round. Also, there are 10 non-overlapping tasks ($T=10$), and for each task, the model is trained for 100 FL rounds. We use Latent Dirichlet Allocation ($\alpha=1$) \cite{reddi2020adaptive} to distribute the data of each task among the clients. We compare the baselines based on three metrics --average accuracy, average forgetting and wallclock time-- which we explain more in the appendix. All the results are reported after averaging over 3 different random seeds. 

\textbf{Baseline.} We compare our method with \textbf{FedAvg} \cite{mcmahan2017communication}, \textbf{FedProx} \cite{li2020federated}, \textbf{FedCIL} \cite{qi2023better}, \textbf{FedLwF-2T}\cite{usmanova2021distillation} and \textbf{Oracle}. \textbf{FedAvg} and \textbf{FedProx} are the two most common aggregation methods in FL. \textbf{FedCIL} is a GANs-based method where clients train the discriminator and generator locally to generate samples from the old tasks. In \textbf{FedLwF-2T}, clients use two teachers -- the global model and their previously trained local model -- to distill their knowledge of the past. Finally, \textbf{Oracle} as an upper bound: during the training of the $i_{th}$ task, clients have access to all of their data from $t=1, ..., i$.

\textbf{Metrics.} We evaluate each approach with the following metrics;
\begin{itemize}
\setlength{\leftmargin}{0pt}
\setlength{\parskip}{0pt}
\setlength{\parsep}{0pt}

\item[--] \textit{Accuracy ($\mathcal{A}^t$)}: Accuracy of the model at the end of task $t$, over all the classes observed so far. 

\item[--] \textit{Average Accuracy ($\Tilde{\mathcal{A}}$)}: Average of all $\mathcal{A}^t$ for all the $T$ available tasks. 
\begin{align}
    \Tilde{\mathcal{A}} = \frac{1}{T} \sum_{t=1}^{T} \mathcal{A}^t
\end{align}

\item[--] \textit{Forgetting (${f}^t$)}: The difference between the highest accuracy of the model on task $t$ and its performance at the end of the training. 

\item[--] \textit{Average Forgetting ($\Tilde{{f}}$)}: Average of the forgetting over all the tasks. 
\begin{align}
    \Tilde{f} = \frac{1}{T-1} \sum_{t=1}^{T-1} f^t
\end{align}

\item[--] \textit{Wallclock time.} This is the time that it takes for the client or server to perform one round of federated learning. The time is measured in seconds and averaged between different clients and rounds. It is worth noting that all the experiments are done in the same GPU, and the number could change by changing the hardware.

\end{itemize}

\subsection{Results}
Table \ref{table:results} shows each method's average forgetting and accuracies. FedAvg and FedProx have the highest forgetting as they are not designed for FCL. Also, high forgetting for FedLwF-2T indicates that extra teachers cannot be effective in the absence of old data. FedCIL and \algname{} have lower forgetting and better accuracy. \algname{} outperforms FedCIL because the generative models in FedCIL need to train for a long time to generate effective synthetic data.

We also compare methods' compute costs. Some methods change after learning the first task; therefore, we distinguish between the cost of the first task and later ones. As depicted, \algname{} can significantly improve accuracy and forgetting at the cost of a slight increase in the clients' training time for $T > 1$ (due to using synthetic data).

The server cost in \algname{} is similar to FedAvg except at the end of each task, where it needs to train the generative model. This extra computation cost should not be a bottleneck because it occurs once per task, and servers usually have access to better computing power compared to clients. 

\section{Discussion}
\subsection{Overheads of generative model} 
\textbf{Client-side.} Using $\gen$ on the client side would increase the computational costs compared to vanilla FedAvg. However, existing methods in CL often need to impose additional costs such as memory, computing, or both to mitigate catastrophic forgetting. Nevertheless, there are ways to reduce costs for MFCL. For example, clients can perform inference once, generate and store synthetic images only for training, and then delete them all. They can further reduce costs by requesting that the server generate synthetic images and send them the data instead of $\gen$. Here, we raise two crucial points about the synthesized data. Firstly, there is an intrinsic distinction between storing synthetic (or $\gen$) and actual data; the former is solely required during training, and clients can delete them right after the training. Conversely, the data in episodic memory should always be saved on the client's side because once deleted, it becomes unavailable. Secondly, synthetic data is shared knowledge that can assist anyone with unbalanced data or no memory in enhancing their model's performance. In contrast, episodic memory can only be used by one client.

\textbf{Server-side.} The server needs to train the $\gen$ \textbf{once per task}. It is commonly assumed that the server has access to more powerful computing power and can compute more information faster than clients. This training step does not have overhead on the client side and, overall, might slow down the whole process. However, tasks do not change rapidly in real life, giving the server ample time to train the generative model before any shifts in trends or client data occur.

\textbf{Communication cost.} Transmitting the generative model can be a potential overhead for \algname{}, as it is a cost that clients must bear \textbf{once per task} to prevent or reduce catastrophic forgetting. However, several possible methods, such as compression, can significantly reduce this cost while still maintaining excellent performance. This could be an interesting direction for future research.

\subsection{Privacy of \algname{}}
Federated Learning, specifically FedAvg, is vulnerable to different attacks, such as data poisoning, model poisoning, backdoor attacks, and gradient inversion attacks \cite{kairouz2021advances,lyu2020threats,fang2020local,geiping2020inverting,chen2022defending,li2020federated}. 

\algname{} generally does not introduce any additional privacy issues and is prone to the same set of attacks as FedAvg. \algname{} trains the generative model based on the weights of the global aggregated model, which is already available to all clients in the case of FedAvg. On the contrary, in some of the prior work, the clients need to share a locally trained generative model or perturbed private data, potentially causing more privacy problems.

For FedAvg, various solutions and defenses, such as differential privacy or secure aggregation \cite{9069945,bonawitz2016practical}, are proposed to mitigate the effect of such privacy attacks. One can employ these solutions in the case of \algname{} as well. Particularly, in \algname{}, the server \textbf{does not} require access to the individual client's updates and uses the aggregated model for training. Therefore, training a generative model is still viable after incorporating these defense mechanisms. 

\algname{} benefits from Batch Statistics Loss ($\loss_{BN}$) in training the generative model. However, some defense mechanisms suggest not sharing local Batch Statistics with the server. While training the generative model without the $\loss_{BN}$ is still possible; it can reduce the accuracy. Addressing this is an interesting future direction.

\section{Conclusion}
This work presents a federated Class-IL framework while addressing resource limitations and privacy challenges. We exploit generative models trained by the server in a data-free fashion, obviating the need for the client's memory.

\nocite{langley00}

\bibliographystyle{icml2023}
\bibliography{ref}

\newpage
\appendix
\onecolumn

\newpage
\appendix
\section{Algorithm in detail}
\begin{algorithm}
    \caption{\algname}
    \label{algo:method}
\begin{algorithmic}[1]
    \STATE $N$: \#Clients, $[\mathcal{C}_{N}]$: Client Set, $K$: \#Clients per Round, $u_i$: client i Update, $E$: Local Epoch
    \STATE $R$: FL Rounds per Task, $T$: \#Tasks, $t$: current task , $|\mathcal{Y}|^t$: Task $t$ Size, $q$: \#Discovered Classes
    \STATE $\model_t$: Global Model for task t, $\gen_t$: Generative Model, $E_{\gen}$: Generator Training Epoch
    \STATE $q \leftarrow 0$
    \STATE $\gen, \model_0, \model_1 \leftarrow \textbf{initialize}()$
    \FOR {$t=1$ {\bfseries to} $T$}
        \STATE $q \leftarrow q + |\mathcal{Y}^t| $
        \STATE $\model_t \leftarrow \textbf{updateArchitecture}(\model_t, q)$
        \FOR {$r=1$ {\bfseries to} $R$}
            \STATE $C_K \leftarrow \textbf{RandomSelect}([\mathcal{C}_{N}], K)$
             \FOR{$c \in C_K$ in parallel}
                 \STATE  $\mathcal{U}_c \leftarrow \textbf{localUpdate}(\model_t, \gen, \model_{t-1}, E)$
            \ENDFOR
        \STATE $\model_t \leftarrow \textbf{globalAggregation}(\model_t, [\mathcal{U}_c])$
        \ENDFOR
        \STATE \# save a frozen version of model for sending to clients
        \STATE $\textbf{saveFrozen}(\model_t)$
        \STATE $\gen \leftarrow \textbf{trainDFGenerator}(\model_t, E_{\gen}, q)$ \#using \eqref{que:gen}
        \STATE $\gen \leftarrow \textbf{freezeModel}(\gen)$ \#fix generator weights 
\ENDFOR
\end{algorithmic}
\end{algorithm}
\subsection{Generative Model Architectures}
In Table \ref{table:architectures}, we show the generative model architectures used for CIFAR-100. The global model has ResNet18 architecture, we change the first $\texttt{CONV}$ layer kernel size to $3\times 3$ from $7\times 7$.
In this table, $\texttt{CONV}$ layers are reported as $\texttt{CONV}K \times K (C_{in}, C_{out})$, where $K$, $C_{in}$ and $C_{out}$ are the size of the kernel, input channel and output channel of the layer, respectively.

    


\subsection{Hyperparameters}
Table \ref{table:param} presents some of the more important parameters.



\begin{minipage}[c]{0.5\textwidth}
\centering
    \begin{tabular}{c}
\hline
 \multicolumn{1}{c}{\textbf{CIFAR-100}} \\
    \hline \hline
    
    \multicolumn{1}{c}{$\texttt{FC}(1000, 128 \times 8 \times 8)$}\\
    \hline
    \multicolumn{1}{c}{$\texttt{reshape}(-, 128, 8, 8)$}\\
    \hline
    \multicolumn{1}{c}{$\texttt{BatchNorm}(128)$}\\ 
    \hline
    \multicolumn{1}{c}{$\texttt{Interpolate}(2)$}\\
    \hline
    \multicolumn{1}{c}{$\texttt{CONV}3\times3(128, 128)$}\\
    \hline
    \multicolumn{1}{c}{$\texttt{BatchNorm}(128)$}\\
    \hline
    \multicolumn{1}{c}{$\texttt{LeakyReLU}$}\\
    \hline
    \multicolumn{1}{c}{$\texttt{Interpolate}(2)$}\\
    \hline
    \multicolumn{1}{c}{$\texttt{CONV}3\times3(128, 64)$}\\
    \hline
    \multicolumn{1}{c}{$\texttt{BatchNorm}(64)$}\\
    \hline
    \multicolumn{1}{c}{$\texttt{LeakyReLU}$}\\
    \hline
    \multicolumn{1}{c}{$\texttt{CONV}3\times3(64, 3)$}\\
    \hline
    \multicolumn{1}{c}{$\texttt{Tanh}$}\\
    \hline
    \multicolumn{1}{c}{$\texttt{BatchNorm}(3)$}

    \end{tabular}
		\captionof{table}{Generative model Architecture}
		\label{table:architectures}

\end{minipage}
\begin{minipage}[c]{0.5\textwidth}
		
		\centering
    \begin{tabular}{c|c}
    \textbf{Dataset} & \textbf{CIFAR-100}\\
    \hline
    \textbf{Data Size} & $32 \times 32$ \\
    \hline 
    \textbf{$\#$ Tasks} & $10$  \\
    \hline
    \textbf{$\#$ Classes\ per task} & $10$\\
    \hline
    \textbf{$\#$ Samples per class} & $500$ \\
    \hline
    \textbf{Batch Size} & 32 \\
    \hline
    \textbf{Synthetic Batch Size} & 32  \\
    \hline
    \textbf{FL round per task} & 100 \\
    \hline
    \textbf{Local epoch} & 10
       \end{tabular}
\captionof{table}{Parameter Settings in different datasets}
\label{table:param}
\end{minipage}





\end{document}